\documentclass{ieeeaccess}
\usepackage{cite}
\usepackage{amsmath,amssymb,amsfonts}
\usepackage{algorithmic}
\usepackage{graphicx}
\usepackage{textcomp}
\usepackage{hyperref}
\usepackage{color,soul}
\usepackage{tabularx}
\usepackage{ragged2e}
\def\BibTeX{{\rm B\kern-.05em{\sc i\kern-.025em b}\kern-.08em
    T\kern-.1667em\lower.7ex\hbox{E}\kern-.125emX}}
\begin{document}
\history{Date of publication xxxx 00, 0000, date of current version xxxx 00, 0000.}
\doi{10.1109/ACCESS.2023.0322000}

\title{Label Propagation Techniques for Artifact Detection in Imbalanced Classes using Photoplethysmogram Signals}
\author{\uppercase{Clara Macabiau}\authorrefmark{1}, 
\uppercase{Thanh-Dung Le}\authorrefmark{1}, \IEEEmembership{Member, IEEE}, \uppercase{K\'evin Albert}\authorrefmark{2}, \uppercase{Mana Shahriari}\authorrefmark{2}, \uppercase{Philippe Jouvet}\authorrefmark{2}, \uppercase{and Rita Noumeir}\authorrefmark{1},
\IEEEmembership{Member, IEEE}}

\address[1]{Biomedical Information Processing Lab, \'{E}cole de Technologie Sup\'{e}rieure, Montr\'{e}al, Qu\'{e}bec, Canada}
\address[2]{CHU Sainte-Justine Research Center, CHU Sainte-Justine Hospital, University of Montreal, Montr\'{e}al, Qu\'{e}bec, Canada}

\tfootnote{This work was supported in part by the Natural Sciences and Engineering Research Council (NSERC), in part by the Fonds de recherche en sant\'e du Qu\'ebec (FRQS).}

\markboth
{Clara Macabiau \headeretal: Label Propagation Techniques for Artifact Detection in Imbalanced Classes using Photoplethysmogram Signals}
{Clara Macabiau \headeretal: Label Propagation Techniques for Artifact Detection in Imbalanced Classes using Photoplethysmogram Signals}

\corresp{Corresponding author: Clara Macabiau (e-mail: clara.macabiau.1@ens.etsmtl.ca).}

\begin{abstract} 
This study aimed to investigate the application of label propagation techniques to propagate labels among photoplethysmogram (PPG) signals, particularly in imbalanced class scenarios and limited data availability scenarios, where clean PPG samples are significantly outnumbered by artifact-contaminated samples. We investigated a dataset comprising PPG recordings from 1571 patients, wherein approximately 82\% of the samples were identified as clean, while the remaining 18\% were contaminated by artifacts. Our research compares the performance of supervised classifiers, such as conventional classifiers and neural networks (Multi-Layer Perceptron (MLP), Transformers, Fully Convolutional Network (FCN)), with the semi-supervised Label Propagation (LP) algorithm for artifact classification in PPG signals. The results indicate that the LP algorithm achieves a precision of 91\%, a recall of 90\%, and an F1 score of 90\% for the "artifacts" class, showcasing its effectiveness in annotating a medical dataset, even in cases where clean samples are rare. Although the K-Nearest Neighbors (KNN) supervised model demonstrated good results with a precision of 89\%, a recall of 95\%, and an F1 score of 92\%, the semi-supervised algorithm excels in artifact detection. In the case of imbalanced and limited pediatric intensive care environment data, the semi-supervised LP algorithm is promising for artifact detection in PPG signals. The results of this study are important for improving the accuracy of PPG-based health monitoring, particularly in situations in which motion artifacts pose challenges to data interpretation.
\end{abstract}


\titlepgskip=-21pt

\maketitle

\section{Introduction}
\label{sec:introduction}
\PARstart{M}{achine} learning, a sub-field of artificial intelligence \cite{helm2020}, has emerged as a transformative technology in various domains, including healthcare. With its ability to analyze large amounts of data \cite{dash2019}, it has the potential to improve healthcare outcomes, help doctors make better decisions \cite{gut2020}, and revolutionize medical research with models that aim to predict injuries \cite{sanchez2016}, detect heart disease earlier \cite{choi2017} and mortality \cite{li2019}. Additionally, machine learning algorithms can contribute to drug discovery and development, optimizing drug efficacy and predicting potential adverse reactions \cite{dara2022}. Machine learning can extract all the necessary information from various types of healthcare data, such as electronic medical records \cite{ho2018}, medical images, and physiological signals. 

Despite its potential, the integration of machine learning into healthcare comes with challenges and considerations. Privacy and ethical implications must be taken into account \cite{ngiam2016}. The data acquired must respect patient privacy and confidentiality and also require standardization and centralized collection for ease of management and consistency, ensuring harmonization \cite{johnson2016}. One major concern is the availability of high-quality data for training and testing these algorithms \cite{hab2016}. To evaluate the performance of the algorithms implemented, it is necessary to have access to a ground truth. Accessing ground truth for evaluating algorithms is challenging, often requiring expert input and large, complete datasets, particularly due to class imbalances in medical data. This further complicates model training, necessitating rebalancing while preserving medical value, with erroneous, missing, or imprecise data exacerbated by artifacts from patient motion or clinical interventions posing additional obstacles to accurate predictions.

During a patient's stay in the hospital, it is important to constantly monitor vital signs. One of these vital signals is the PPG signal, which is frequently captured during different types of movements, introducing motion noise and interfering with the accuracy of the signals. This noise is irregular and causes high-amplitude fluctuations within the PPG signals \cite{poll2022}. Motion artifacts can result in the pulse oximeter either misinterpreting movement as the actual signal or masking the true signal with unwanted interference, leading to incorrect readings, false alarms, and missed important alarms \cite{pet2007}. The main objective of this work is to detect motion artifacts in PPG signals obtained from the Pediatric Intensive Care Unit (PICU) database of the CHU Sainte-Justine Hospital (CHUSJ). The cleaned PPG signals will be used to construct clinical decision systems (CDSS) at CHUSJ's PICU. Specifically, annotated signals will be used in screening and identifying various health-related concerns in children. For example, changes in blood pressure in children are significant indicators for identifying patients who require immediate care and admission to the PICU. Invasive methods, like catheter insertion for continuous blood pressure monitoring, offer precise real-time data but come with significant risks such as bleeding and infection \cite{kim2014}. On the other hand, conventional cuff-based measurements, though less invasive, provide only intermittent readings and may not capture sudden clinical changes effectively. Therefore, predicting blood pressure from PPG waveforms has emerged as a successful approach \cite{hill2021} for comprehensive CDSS applications.

This study contributes to the field in three main ways. Firstly, we compare resampling methods commonly used in medical data analysis to address the imbalance between clean PPG samples and artifact-contaminated ones. Secondly, we validate the efficacy of the LP algorithm for motion artifact detection within PPG signals, offering insights into its performance in scenarios with limited labeled data. Lastly, we present a detailed performance comparison between traditional supervised algorithms and the semi-supervised LP approach, highlighting the advantages of leveraging unlabeled data in artifact classification tasks.

\section{Related work}

Numerous methods have already been developed to detect motion artifacts in PPG signals. First, the traditional methods are easy to implement. In \cite{hanyu2017}, the authors used statistical analysis to compare the values of three statistics calculated for each pulse of the PPG signal to determine which pulses are noisy. This method will be used in the labeling step for the rest of the project. Adaptive filtering is another method of artifact detection \cite{wu2017}. The adaptive filter uses an algorithm that continuously updates its coefficients to obtain an error signal as close as possible to the original PPG signal. Both approaches have the advantage of being easy to implement but are notably sensitive to empirical thresholds. Among other popular methods, the wavelet transform uses cascaded high-pass and low-pass filters to obtain the desired signal decomposition. Once the signal has been decomposed, the coefficients are analyzed to identify any artifacts \cite{joseph2014}. Empirical mode decomposition, like the wavelet transform, is a time-frequency analysis of the signal \cite{wang2010}. When these modes are obtained, the objective is to calculate the instantaneous frequency for each mode to detect modes that have a frequency close to the harmonics of a PPG signal and modes characteristic of motion artifacts. So, these methods have the advantage of being fast and simple, which is useful, but when used alone, they have limited adaptability and may not work as well with complex movements or unexpected scenarios. A summary table with the review activity is presented in Table \ref{tab:review}. Therefore, we decided to use a combination of signal processing algorithms for the preprocessing part and machine learning models.

Regarding machine learning models, in \cite{xu2015}, the authors explore using semi-supervised models to classify temporal data. These models are based on a graphical approach like the LP algorithm. The algorithm's results are evaluated on different datasets of varying lengths, including ECG (electrocardiogram) signal data. The results show that semi-supervised models are accurate for classifying time series data. However, these algorithms have not been applied to artifact detection. Semi-supervised learning is widely used as a classification algorithm in cases where not all data is annotated. Active learning is also a powerful semi-supervised classification method that has proven effective for temporal data \cite{shin2021coherence}. In our scenario, the LP algorithm is effective because the availability of labeled data is limited, and there is a large amount of unlabeled data \cite{bunger2022}. Considering this information, the LP algorithm was implemented for this project.

In the first part of the project, the LP algorithm is used for data annotation. First, an expert annotated a small proportion of data, and a statistical analysis algorithm was used to validate the annotations. Then, the LP annotates all data using only a small proportion of previously annotated data. Our medical data are unbalanced, with around 80\% of pulses free of artifacts and only 20\% with artifacts. This means that to have an accurate labeling algorithm, a rebalancing of the classes in the training part needs to be done. Several methods are available for this: oversampling, undersampling, and both oversampling and undersampling. It must be remembered that medical data is being worked with, so sampling methods must make medical sense, whether by randomly duplicating data or by removing it. Medical data involves intricate relationships among data elements, such as patient demographics, medical history, symptoms, diagnoses, treatments, and outcomes \cite{mazu2008}.

\begin{table*}[t]
\centering
\caption{Summary table of the literature review}
\label{tab:review}
\begin{tabularx}{\textwidth}{|>{\RaggedRight}X|>{\RaggedRight}X|c|>{\RaggedRight}X|>{\RaggedRight\arraybackslash}X|}
\hline
Authors &  Title & Year & Source & Findings\\ \hline
Q. Wang et al. & Artifact reduction based on empirical mode decomposition (EMD) in photoplethysmography for pulse rate detection & 2010 & 2010 Annual International Conference of the IEEE Engineering in Medicine and Biology & Empirical Mode Decomposition and Hilbert transform combined give good results for the decomposition of PPG signals and the reduction of motion artifacts\\ \hline
G. Joseph et al. & Photoplethysmogram (PPG) signal analysis and wavelet de-noising & 2014 & 2014 Annual International Conference on Emerging Research Areas: Magnetics, Machines and Drives (AICERA/iCMMD) & Unwanted PPG signal interference is successfully removed using wavelet transform while preserving the signal information \\ \hline
Z. Xu and K. Funaya & Time series analysis with graph-based semi-supervised learning & 2015 & 2015 IEEE International Conference on Data Science and Advanced Analytics (DSAA) & The use of a new probabilistic semi-supervised method combining different graph constructions and distance techniques gives better results on different types of real data \\ \hline
S. Hanyu and C. Xiaohui & Motion artifact detection and reduction in PPG signals based on statistics analysis & 2017 & 29th Chinese Control And Decision Conference (CCDC) & The use of statistical thresholds on corrupted PPG segments correlated with high-quality segments effectively removes motion artifacts\\ \hline
C.-C. Wu et al. & An implementation of motion artifacts elimination for PPG signal processing based on recursive least squares adaptive filter & 2017 & 2017 IEEE Biomedical Circuits and Systems Conference (BioCAS) & Adaptive approach to remove motion artifacts, using DC Remover method and Recursive Least Squares adaptive filter\\ \hline
Y. Shin et al. & Coherence-based label propagation over time series for accelerated active learning & 2021 & International Conference on Learning Representations & A new active learning method for time series, called TCLP, improves classification accuracy when a very small number of data points are already labeled \\ \hline
D. Bünger et al. & An empirical study of graph-based approaches for semi-supervised time series classification & 2022 & Frontiers in Applied Mathematics and Statistics, vol. 7 & Comparison of different distance measures in the implementation of graph-based models, including some semi-supervised models, in classifying binary time series datasets  \\ \hline
\end{tabularx}
\end{table*}

Another aim of this project is to compare classifiers to the LP algorithm, used as a classifier, to accurately detect artifacts. In health care, classifiers are a real help in decision-making \cite{robbins2011}. The spectrum of classifiers is very wide: from traditional classifiers like KNN, Support Vector Machine (SVM), Decision Tree (DT), and Naive Bayes classifier (NB) \cite{al2015}, to classifiers using neural networks, such as MLP or Transformers. A comparison of the results of each type of classifier with the semi-supervised LP algorithm will be presented. The effectiveness of these two streams is analyzed by the experimental results (in section \ref{sec:results}) from the comparative analysis of semi-supervised LP (with KNN kernel) and fully-supervised learning, including conventional machine learning classifiers (KNN, Support Vector Classification (SVC), DT, Random Forest (RF), GaussianNB, MultinominalNB, and Logistic Regression (LR)), MLP, and Transformers. Then, the best classification method will be presented, followed by a conclusion on artifact detection.

The paper is structured as follows. In section \ref{sec:methode}, data characteristics, preprocessing, methodology, labeling, and classification are introduced. In section \ref{sec:experience}, the implementation of experiments is presented. Section \ref{sec:results} is used to evaluate the results with different metrics and present a comparison of experimental result tables. In section \ref{sec:conclusion}, the results are interpreted, and the limitations are discussed.

\section{Material and Methods}
\label{sec:methode}
\begin{figure}
	\centering
	\vspace{2pt}
	\includegraphics[scale=0.6]{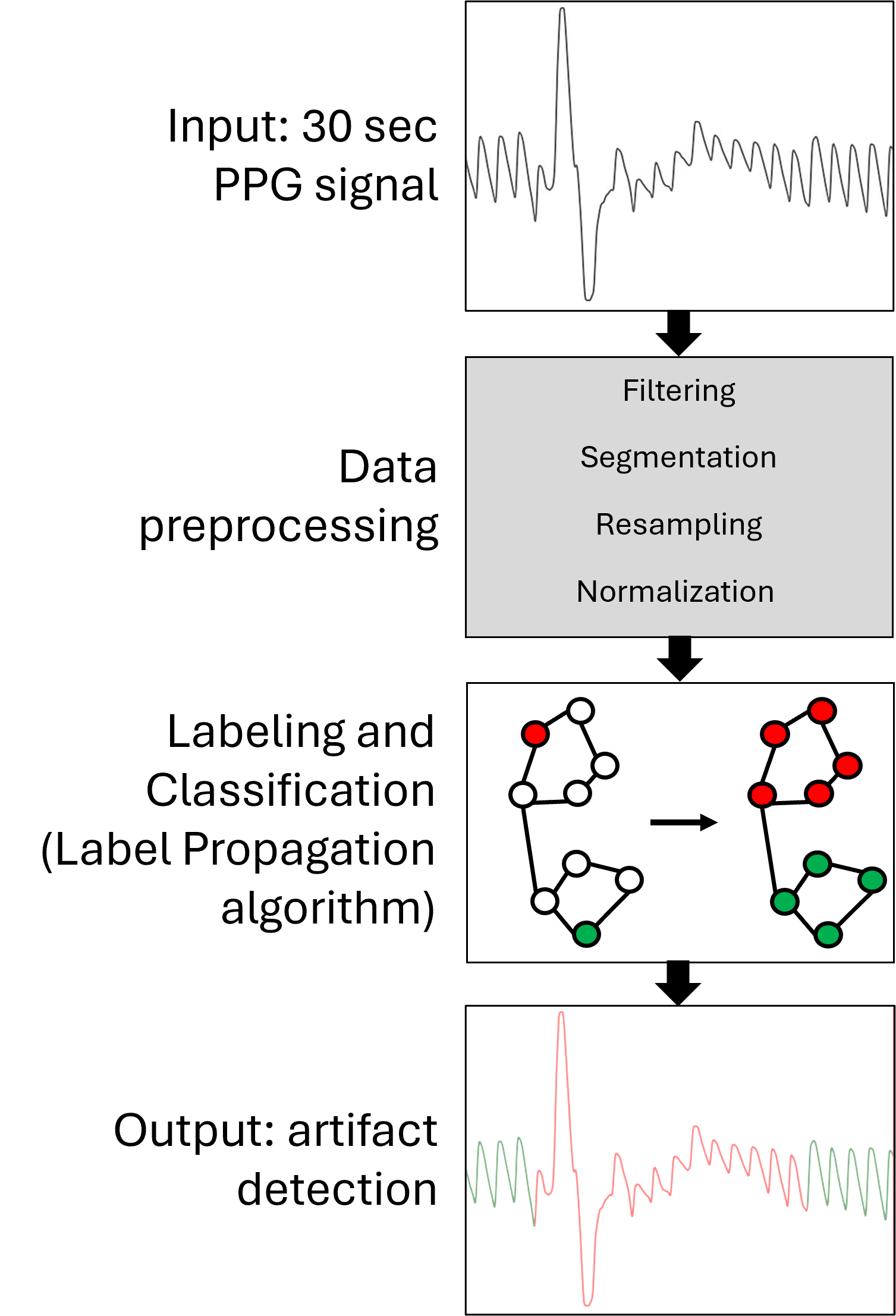}
	\caption{Workflow of the proposed method for detecting motion artifacts in PPG signals.}
	\vspace{-8pt}
	\label{fig:workflow}
\end{figure}

This study was conducted following ethical approval from the research ethics board at CHUSJ (protocol number 2023-4556, accepted January 18, 2023). The detailed workflow of the various work stages is shown in Fig. \ref{fig:workflow}. Specifically, the workflow of the proposed method for detecting motion artifacts in PPG signals begins with the input of a 30-second PPG signal. This signal undergoes data preprocessing, which includes filtering, segmentation, resampling, and normalization to prepare the data for analysis. Next, a labeling and classification step is performed using a label propagation algorithm to identify and classify segments of the signal. Finally, the process outputs artifact detection, highlighting the portions of the signal affected by motion artifacts.

\subsection{Data Collection}
This project aims to detect motion artifacts in PPG signals. The eligible study population includes all children aged 0 to 18 years, admitted between September 2018 and July 2022 inclusive, for whom electrocardiogram (ECG), PPG, and arterial blood pressure (ABP) waveform records are available. In this population, specific exclusion criteria have been established to avoid bias. Data collected beyond the fourth day of hospital stay will be disregarded to prevent potential bias from a few patients who may have prolonged stays with arterial lines. Patients on extracorporeal membrane oxygenation (ECMO) treatment will also be excluded from the analysis. Furthermore, if a patient is readmitted to the PICU multiple times, only data from the first stay will be analyzed.

A PPG signal is recorded using a sensor called the pulse oximeter. This device is placed on a patient's skin, for children, on a fingertip or earlobe. A PPG sensor emits light into the skin, partially absorbed by the blood vessels. Changes in blood flow during the cardiac cycle cause variations in light absorption. The sensor detects the reflected light, measuring its intensity modulated by blood volume changes. This varying intensity is converted into an electrical signal, creating the PPG waveform. Blood pressure signals are recorded using an invasive and continuous method, i.e., the catheter, and a non-invasive and discontinuous method, i.e., the blood pressure cuff. ECG is continuously recorded by placing electrodes on the patient's chest.
The Sainte-Justine University Hospital PICU utilizes a high-resolution research database (HRDB) \cite{brossier2018,roumeliotis2018} that has been approved by the ethical committee. The HRDB links biomedical signals extracted from the different devices, displayed through patient monitors, to the electronic patient record continuously throughout their stay in the unit \cite{mathieu2021}.

Between 2018 and 2022, 1571 patients met the inclusion criteria. For each patient, four physiological signals were extracted: ECG, PPG, blood pressure from the catheter, and blood pressure from the cuff. Each signal was extracted over 96 hours (4 days). Signal values are grouped together in a table with the date and time of acquisition. For the PPG signal, 640 values are acquired every 5 seconds, corresponding to a sampling frequency of 128 Hz. For blood pressure and ECG signals, 2560 values are acquired every 5 seconds, with a sampling frequency of 512 Hz.  For the duration of the extraction, a fixed 30-second window of PPG signals will be used for further processing.

\subsection{Preprocessing}
The raw PPG signal is preprocessed to increase its quality, remove unwanted noise, and make it more suitable for subsequent processing steps \cite{lim2018}. The different steps are described below:
\begin{enumerate}
    \item \textbf{Filtering}: each signal window is filtered using a band-pass Butterworth filter; the cut-off frequencies are 0.5 and 5 Hz, corresponding to a heart rate between 30 and 300 bpm. A forward-backward filtering is used to avoid phase distortions. The objective is to remove baseline wander and high-frequency noise. 
    \item \textbf{Pulse segmentation}: a function to find all local minima by comparing samples is used. The aim is to divide the preprocessed PPG signal into smaller segments or windows to detect the artifacts present for each pulse. In our case, a segment is a pulse. The size of each segment may vary depending on the characteristics of the PPG signal and the specific application of the signal pulses. A pulse is considered to lie between two minima. 
    \item \textbf{Resampling}: the duration of a cardiac cycle for children is between 0.3 and 1 second. A pulse represents a cardiac cycle. Therefore, not all pulses have the same number of samples. Each pulse is uniformly oversampled in time to contain 256 samples, corresponding to a heart cycle of 1s. A linear interpolation function \cite{li2012} is used to create the missing points for each pulse. Linear interpolation is favored for signals due to its simplicity, computational efficiency, and ability to estimate values between known data points. It maintains signal continuity and linearity, making it suitable for signals with relatively smooth and linear variations.
    \item \textbf{Normalization}: the data are normalized to have a unit variance and zero mean. This normalization ensures that all features or variables in the data have the same scale, preventing certain features from dominating the learning process simply because they have larger numerical values.
    \item \textbf{Data transformation}: 
    each PPG pulse, essentially a waveform representing blood volume changes over time, can be represented as a data point in a column containing 256 values. These values are equally spaced points obtained using step 3 of the preprocessing. At the end of preprocessing, a vector of 256 points is obtained, representing a pulse of the PPG signal. The number of vectors depends on the number of pulses. This method allows us to work with PPG data in a structured manner suitable for various applications, from statistical analysis to machine learning.
\end{enumerate}
\begin{figure}[b]
	\centering
	\vspace{2pt}
	\includegraphics[scale=0.6]{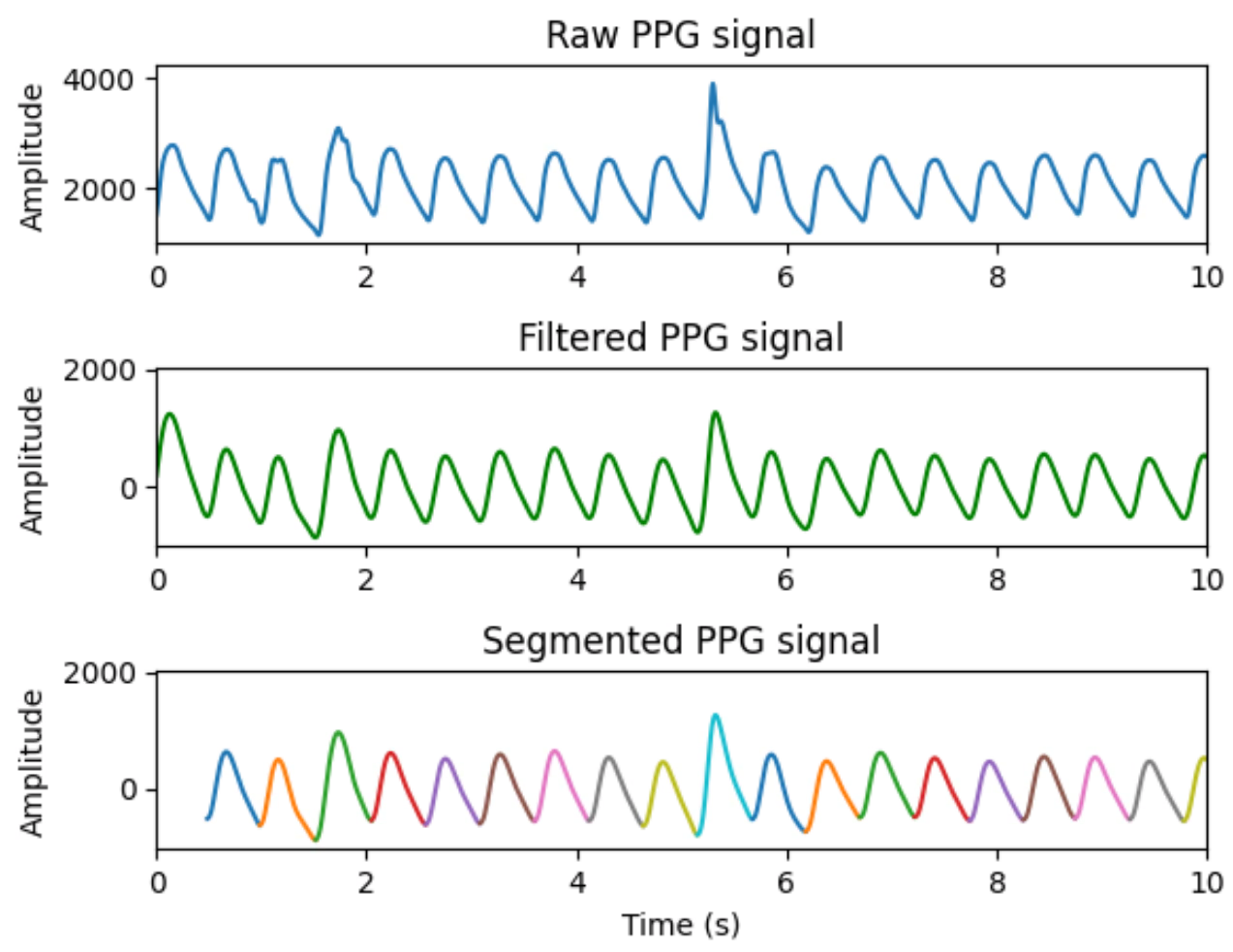}
	\caption{Example of a 10s segment of a 30s raw PPG signal in the top image, filtered signal in the middle image, and segmented signal in the bottom image.}
	\vspace{-8pt}
	\label{fig:pre}
\end{figure}
Fig. \ref{fig:pre} shows the first 10 seconds of a raw PPG signal, when the signal has been filtered, and finally when the pulses have been segmented. The effect of the bandpass filter can be seen in the second figure. The filter has smoothed the signal by removing the extreme frequency components. The signal waveform is preserved, and the filter does not introduce resonances or significant ripples in the desired frequency range. Note that the first pulse has not been segmented. This is because the function could not detect the two minima that make up a pulse and, therefore, could not segment it. The signal does not start at the first low point of the pulse.

\begin{figure}[b]
	\centering
	\vspace{2pt}
	\includegraphics[scale=0.8]{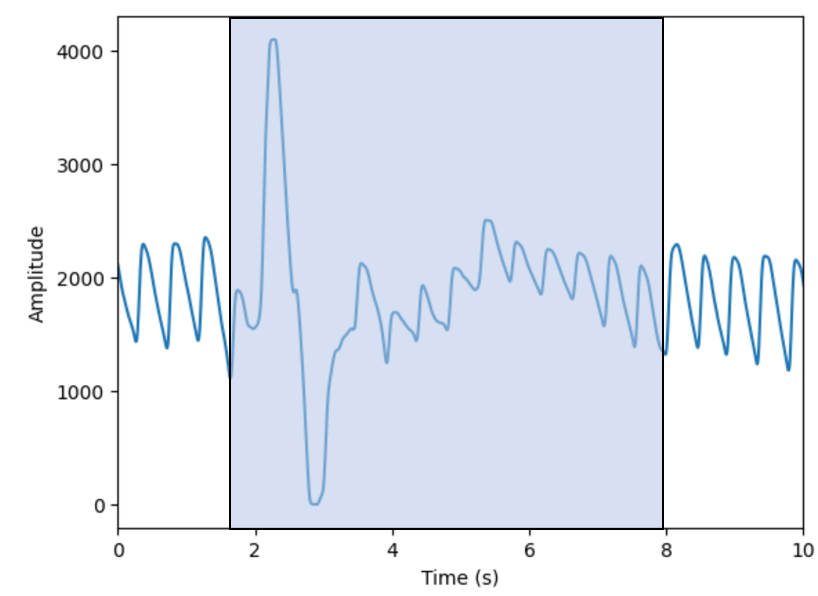}
	\caption{Example of a 10s segment of a 30s raw PPG signal. Inside the blue box are all the pulses containing motion artifacts. }
	\vspace{-8pt}
	\label{fig:demo}
\end{figure} 

\subsection{Dataset annotation}
First, after preprocessing the data, the aim is to build a ground truth for future evaluation of the classification algorithms. To this end, one human expert annotated 10\%  of the database. Annotation is visual, comparing pulses with each other and binary classifying each pulse as good or as containing artifacts. 
To avoid involving another human expert, in consideration of time and specialist resources, we implemented an automated algorithm to handle additional annotations. This algorithm, acting as a surrogate expert, was developed to recheck the entirety of the 10\% annotated data by the human expert, identifying similarities in the process. Employing a statistical approach, the algorithm determines if the values of a given pulse lie within standard parameters or deviate from the norm. We subsequently cross-validated the algorithm's annotations against those from the professional expert to determine the algorithm's accuracy. It was decided to annotate a maximum of 10\% of the database and then use the LP algorithm to annotate the rest of the data.

\subsubsection{Expert labeling}
PPG signals already segmented are presented to the expert. By analyzing each pulse, the expert classifies each pulse as artifact or artifact-free. A pulse is defined as artifact-free if its morphology is typical, i.e., if its characteristics - amplitude, width, shape - are the same as those of adjacent signals. A pulse is defined with artifacts if its characteristics differ from those of adjacent pulses (see fig. \ref{fig:demo}). To recheck the annotations of one expert, an algorithm that acts as a second expert was set up, allowing all impulses to be reannotated to see similarities. This algorithm uses a statistical approach to assess whether the statistical values of a pulse are normal or outside the norm.

\subsubsection{Statistical analysis}
\label{subsec:stat}
For each cardiac cycle, which corresponds to a pulse, if the waveform is similar, statistics such as skewness, standard deviation (std), and kurtosis are approximately constant for each cycle. It is, therefore, possible to detect motion artifacts by using the value of these statistics to differentiate a pulse without artifacts from a pulse with motion artifacts \cite{hanyu2017}. Skewness indicates the degree of asymmetry in the probability distribution of a random variable around its mean. It can take on positive, zero, negative, or undefined values, reflecting the shape and symmetry of the distribution. Kurtosis is the sharpened of the peak of a frequency-distribution curve, and standard deviation reflects the dispersion degree of a data set. If $X$ is considered a variable with $\mu$ and $\sigma$, the mean and standard deviation, respectively, statistical values are calculated as follows: 
\begin{equation}
    \text{Kurt}[X] = E\left[\left(\frac{X-\mu}{\sigma}\right)^4\right] = \frac{E\left[\left(X-\mu\right)^4\right]}{\sigma^4}
\end{equation}
\begin{equation}
    \text{Skew}[X] = E\left[\left(\frac{X-\mu}{\sigma}\right)^3\right] = \frac{E\left[\left(X-\mu\right)^3\right]}{\sigma^3}
\end{equation}
\begin{equation}
    \text{std}[X] = \sqrt{E\left[\left(X-\mu\right)^2\right]}
\end{equation}
These values are calculated for each pulse of a signal. So, in our case, the variable X represents a vector of all the samples in a pulse. If the shape of the cycle changes, then these statistical values will no longer be constant. To be able to detect outliers, thresholds that detect skewness, kurtosis, and standard deviation values that are not normal, i.e. values for artifact-free cycles, were set up. For this reason, the distribution of each of these three statistics over a pulse can be estimated using a normal distribution \cite{kris2008}. The aim is to reduce the risk of a pulse being incorrectly annotated. To do this, a wide confidence interval is taken to ensure that the probability that the value of the corresponding statistic is not unnecessarily rejected. If X is considered to be a variable that can be approximated by a normal distribution $\mathcal{N}(\mu,\,\sigma^{2})$, the probability that this variable lies within the chosen confidence interval can be written as follows:
\begin{equation}
    \mathbb{P}(\mu-2\sigma \leq X \leq \mu+2\sigma) \approx 0.9545
\end{equation}
After several experiments, this 95\% confidence interval gives the best results, as it reduces the risk of poor detection. So, lower and upper thresholds can be defined as follows: 
\begin{equation}
    th_l = \mu-2\sigma
\end{equation}
\begin{equation}
    th_u = \mu+2\sigma
\end{equation}
The mean and standard deviation are calculated for each statistic measured by taking the set of values for each pulse of a signal. A waveform segment is classified as containing motion artifacts to effectively detect motion artifacts if at least one of the three statistics falls outside the defined thresholds. The result of this first step is a small proportion of the annotated dataset, with a binary value for each pulse: pulse with artifact or without artifact. The annotations given by the algorithm are then compared with the expert's annotations and found to have 80\% similarity. After examining the annotations with the expert, the function chosen to segment the pulses did not always correctly segment a pulse that was formed by a distinct diastole and systole curve. In a cardiac cycle, diastole is the relaxation phase when the heart fills with blood, and systole is the contraction phase when the heart pumps blood out to the body or lungs. In this case, two pulses were detected instead of one. This segmentation error partly explains the 20\% difference in annotation between the expert and the algorithm. The percentage of similarity is considered high enough to validate the expert's annotations.

\subsubsection{Imbalanced dataset}
\label{subsec:imbalanced}
The two classes of annotated data are unevenly distributed. The annotation includes many more pulses without motion artifacts, approximately 80\% and 20\% of pulses with motion artifacts. For accurate results with the algorithms, the data needs to be resampled. The complex characteristics of our clinical data, such as small training sizes, many features, and correlations between the features, make the task more complicated. Understanding the interconnectedness of these variables is crucial for accurate analysis and prediction \cite{mazu2008}. 
Oversampling and undersampling methods are the most frequently used. Under-sampling reduces the majority of class examples, achieving a balanced dataset, with random under-sampling (RUS) being a well-known method. However, under-sampling may lead to the loss of valuable information from the majority class. On the other hand, over-sampling increases the minority class examples. Random over-sampling (ROS) replicates existing minority examples, but it may result in overfitting. Synthetic minority over-sampling technique (SMOTE) generates artificial minority examples by interpolating between selected examples and their nearest neighbors. Modifications such as adaptive synthetic sampling (ADASYN) adjust the number of artificial minority examples based on the density of majority examples surrounding the original minority example \cite{fuji2020}. Also, it is concluded that there is no clear winner between oversampling and undersampling to compensate for the class imbalance if factors such as class distribution, class prevalence, and features correlations in medical decision-making \cite{mazu2008} are not taken into consideration. In the section \ref{sec:results}, the different results obtained with the sampling methods will be presented to conclude on the best method for our study.

\subsubsection{Label Propagation}
The LP algorithm is an iterative algorithm that assigns labels to unlabeled data points by propagating labels through the dataset. It was first presented in an article published in 2002 by X. Zhu and Z. Ghahramani, entitled "Learning from labeled and unlabeled data
with label propagation" \cite{zhu2002}. In graph-based semi-supervised learning methods, a graph where each node is represented by a vector of features is created. The edges between nodes are weighted based on how similar the features are. When the weights of the edges are high, it means that the connected nodes are likely to have the same label. This idea is based on the assumption that samples close to each other in the graph are part of the same group or category \cite{song2022}. At the start of the algorithm, only a small proportion of the data is already labeled, corresponding here to the proportion of data annotated in the previous step. In our case, considering that we have around 51 pulses per signal and that we have annotated 10\% of the entire database of 1571 signals, we therefore have 8000 pulses, and thus 8000 nodes in the graph. This algorithm is based on the hypothesis that if two nodes are connected, they carry a similarity. Usually, the Euclidean distance between nodes is calculated to establish the graph. Depending on the kernels chosen for the algorithm's operation, this distance measurement may be different. Consider the following notations: 
\begin{align*}
    u &: \text{number of unlabeled points}\\
    l &: \text{number of labeled points}\\
    k &: \text{number of classes}
\end{align*}
In the final state, this algorithm aims to look at all the probabilities a node has of belonging to a certain class and take the largest. $Y$ a matrix with rows containing the probabilities that a node belongs to a certain class is considered. This matrix $Y$ is a $N\times k$ matrix where $N=l+u$. Also considered $T$, a $N\times N$ probability transition matrix. This matrix T is obtained by calculating the degree matrix ($D$) and the adjacency matrix ($A$). It defines the probability of jumping from one node to another in $t$ steps. This number $t$ can tend towards infinity \cite{bodo2015}. The matrix $Y$ contains two sub-matrices: $Y_l$ and $Y_u$, respectively, for the known and unknown labels. The same applies to the $T$ matrix, which contains 4 sub-matrices: 
\begin{itemize}
    \item $T_{ll}$: probability to get from labeled nodes to labeled nodes. This matrix will be an identity matrix. 
    \item $T_{lu}$: probability of getting from labeled to unlabelled nodes. This will be a zero matrix because labelled nodes are absorbing states, it means you are in a self-loop and can't move in any direction.
    \item $T_{ul}$ and $T_{uu}$: probability to get from unlabelled nodes to labeled and unlabelled nodes, respectively. 
\end{itemize}
Consider $\hat{Y}$, the probability matrix of annotations obtained in the final state. The matrix T is set to the infinite power, and $Y_0$ represents the initial annotations of the nodes. The equation for the final stage of this algorithm can be expressed as: 
\begin{equation}
\label{eqn:lp}
    \hat{Y} = T^{t \rightarrow \infty}Y_0
\end{equation}

\begin{figure*}[t]
	\centering
	\vspace{2pt}
	\includegraphics[scale=0.4]{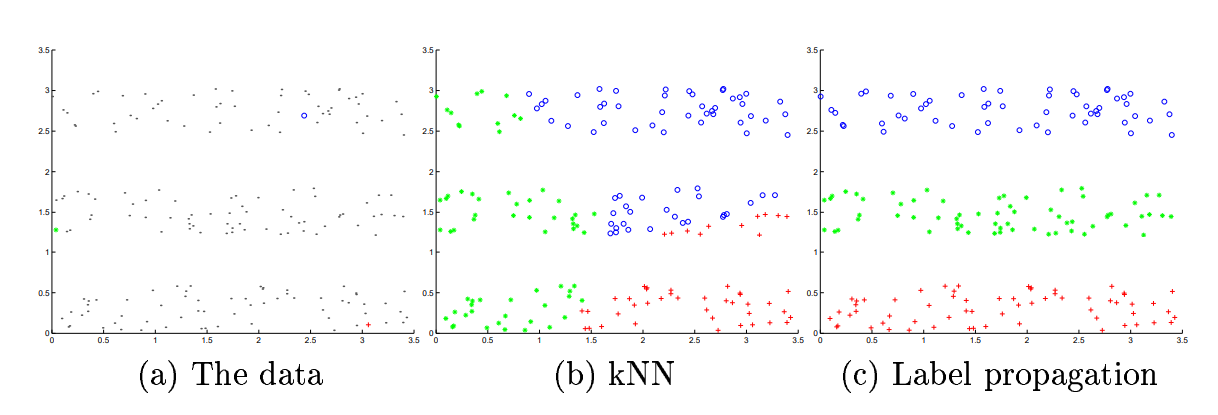}
	\caption{Comparison of label propagation between KNN and LP model with the "3 Bands dataset". (a) 3 initial annotated points (3 classes represented in green, red, and blue) and 178 non-annotated points (b) annotated dataset with KNN (c) with LP. From \cite{zhu2002}.}
	\vspace{-8pt}
	\label{fig:lp_original}
\end{figure*}

In \ref{eqn:lp}, the matrix $T$ is set to the power $t$ with $t$ tending to infinity. It can be written as: 
\begin{equation}
    \lim\limits_{t \to \infty} T^t = 
    \begin{bmatrix}
    I & 0\\
    \left ( \sum_{t=0}^{\infty}T_{uu}^t  \right )T_{ul} & T_{uu}^{\infty}
    \end{bmatrix}
\end{equation}
The sum between the brackets is similar, when $t$ tends to infinity, to a geometric series that has an argument that is less than 1 in modulus. And if $T_{uu}$ is multiplied by itself a large number of times, knowing that the values are less than 1, it will become very close to 0. Therefore, a conclusion on the limit of the transition matrix for a very large number of steps is: 
\begin{equation}
    \lim\limits_{t \to \infty} T^t = 
    \begin{bmatrix}
    I & 0\\
    \left ( I-T_{uu}  \right )^{-1}T_{ul} & 0
    \end{bmatrix}
\end{equation}
The equation \ref{eqn:lp} can, therefore, be rewritten: 
\begin{equation}
\begin{bmatrix}
\hat{Y_l} \\
\hat{Y_u}
\end{bmatrix} = 
\begin{bmatrix}
I & 0\\
\left ( I-T_{uu}  \right )^{-1}T_{ul} & 0
\end{bmatrix} 
\begin{bmatrix}
Y_{l0} \\
Y_{u0}
\end{bmatrix}
\end{equation}
For unknown labels, the following formula can be written: 
\begin{equation}
\label{eqn:lpfinal}
    \hat{Y_u} = \left ( I-T_{uu}  \right )^{-1}T_{ul}Y_{l0}
\end{equation}
This matrix contains the new labels and is the output of the algorithm. To sum up, the various stages of the algorithm can be summarized as follows:
\begin{enumerate}
    \item Creation of a graph with nodes labeled and unlabeled.
    \item Calculation of the probability transition matrix $T$. This matrix is linked to the degree matrix $D$, a diagonal matrix where each diagonal element corresponds to the sum of edge weights connected to that node. Also linked to the adjacency matrix $A$, it is a square matrix where each row and column corresponds to a node, and the value at the intersection indicates whether there's an edge (value 1, otherwise 0) connecting those nodes. The formula is: $T = D^{-1}\cdot A$. This matrix is the same throughout the algorithm.
    \item Calculation of the new labels for each $t$ iteration: 
    \begin{equation}
        Y^{t+1} = T^tY^t
    \end{equation}
    \item Repeat step 3 until convergence.  
\end{enumerate}
A concrete example of how the LP algorithm works is shown in figure \ref{fig:lp_original}. This example is based on a sample of synthetic data where each of the three classes is represented in a band. The KNN algorithm is unaware of the band structure and fails to propagate labels efficiently. The LP model, on the other hand, recognizes this structure and uses it to its advantage to group labels.

While performing label propagation, groups of closely linked nodes quickly reach a consensus on a single label, causing many labels to vanish. Only a few labels remain after propagation. When nodes end up with the same label after convergence, it signifies that they are part of the same group.

\subsection{Classification}
\label{subsec:classification}
Once the ground truth has been established, the aim is to classify the pulses and compare the results obtained with the annotations. Machine learning classifiers are used for classification. These automatic algorithms categorize data into the two classes of our problem. They operate as mathematical models, utilizing statistical analysis and optimization techniques to detect patterns within the data. By identifying these patterns, classifiers can assign each instance to a specific class or category. There are a wide variety of traditional classifiers, both supervised and unsupervised. Supervised classifiers have been chosen to be utilized to process medical data, which is also temporal data. Here are 4 examples \cite{al2015}: 
\begin{enumerate}
    \item \textbf{KNN}: this is a supervised method where k represents the number of neighbors. For classification, when given a new input data point, the algorithm identifies the k nearest neighbors from the training dataset based on their feature similarity. The class label of the majority of these k neighbors is then assigned to the new data point.
    \item \textbf{SVM}: in SVM, data points are mapped as vectors within a high-dimensional space. The algorithm aims to identify the optimal hyperplane that distinctly categorizes the classes. In binary classification, a hyperplane can be considered as a boundary delineating two distinct data classes. While numerous hyperplanes might achieve this separation, the algorithm selects the one that provides the most effective separation. For a specific classification purpose, as is the case for this project, we have subsequently used SVC, a type of SVM specialized for classifications. 
    
    \item \textbf{DT}: Each internal node represents a feature or attribute, and each branch represents a decision rule based on that feature. The leaf nodes of the tree represent the final class label or predicted value. During training, each value is separated based on the attribute. When making predictions, new data points traverse the decision tree by following the decision rules at each node until reaching a leaf node, which then provides the predicted class label or value.
    \item \textbf{NB classifier}: this classifier uses probability to predict whether an input will fit into a certain category. It builds a statistical model based on these probabilities. Naive Bayes calculates the likelihood of the data point belonging to each class using the previously estimated probabilities.
\end{enumerate}
Traditional classifiers have big advantages for small or medium datasets that require simpler or linear models. They have few layers in their architecture; conversely, deep learning (MLP and Transformers) architectures comprise multiple layers of neural networks. Deep architectures take advantage of unsupervised pre-training at the layer level, which facilitates efficient tuning of the deep networks and enables them to extract intricate structures from input data. These extracted features at higher levels contribute to improved predictions and overall performance \cite{miotto2018}. For classification, MLP and Transformers are neural networks classifiers: 
\begin{itemize}
    \item \textbf{MLP}: it consists of multiple layers of nodes (neurons) that are interconnected through weighted connections. MLP employs a feedforward mechanism, where information flows from the input layer through the hidden layers to the output layer. Each node in the network applies an activation function to the weighted sum of its inputs to produce an output. Through a process called backpropagation, the MLP classifier adjusts the weights to minimize the error between predicted and actual labels during training.
    \item \textbf{Transformers}: it relies on the attention mechanism. The attention-mechanism looks at an input sequence and decides at each step which other parts of the sequence are important. A Transformer is an architecture for transforming one sequence into another one with the help of two parts (Encoder and Decoder).
\end{itemize}
The objective is to apply all these classifiers to the PPG signal pulses so that a comparison of the classifiers on our medical data can be built. In addition to being compared with each other, these classifiers will also be compared with the LP semi-supervised algorithm, which annotates our database and classifies artifacts in PPG signals.

\section{Experimental Implementation}
\label{sec:experience}
First, as a reminder, in the LP algorithm, the two input matrices are the annotation matrix, a binary vector, and a matrix containing the features for each pulse. Each pulse represents a node in the algorithm's graph. For the choice of features, the signal from a temporal perspective has been considered. Therefore, an input matrix for the algorithm of size 256 samples $\times$ the number of pulses can be obtained.

Different metrics have been chosen to evaluate our results. The negative state (0) is a pulse without motion artifacts, whereas the positive state (1) is a pulse with motion artifacts. All these measures are based on the evaluation of false negatives (FN), pulse with artifact incorrectly identified as a clean pulse, false positives (FP), clean pulse incorrectly identified as a pulse with artifact, true negatives (TN), clean pulse correctly identified as a clean pulse, and true positives (TP), pulse with artifact correctly identified as a pulse with artifact. The following metrics are defined: 
\begin{itemize}
    \item \textbf{Confusion Matrix}: a table with two rows and two columns that reports the number of true positives, false negatives, false positives, and true negatives.
    \item \textbf{Precision, Recall, and F1}: these three scores give a more general idea of how the algorithm works, rather than just looking at the algorithm's accuracy, which can be biased in certain situations. They are defined as follows: \begin{equation*}
        \text{Precision} = \frac{TP}{TP+FP}
    \end{equation*}
    \begin{equation*}
        \text{Recall} = \frac{TP}{TP+FN}
    \end{equation*}
    \begin{equation*}
        \text{F1} = 2 \times \frac{\text{Precision} \times \text{Recall}}{\text{Precision}+\text{Recall}}
    \end{equation*}
    \item \textbf{Matthews Correlation Coefficient (MCC), Cohen's Kappa Coefficient, and Critical Success Index (CSI)}: MCC is particularly useful in the case of binary classification, where the two classes are unbalanced. It varies between 0 and 1. CSI is also known as the Threat Score. A CSI of 1 indicates perfect prediction, while a score of 0 indicates no successful predictions beyond random chance. Kappa Coefficient ($\kappa$) is stronger than accuracy; it ranges from -1 to 1.
    \begin{equation*}
        \text{mcc} = \frac{TP \times TN - FP \times FN}{\sqrt{(TP + FP)(TP + FN)(TN + FP)(TN + FN)}}
    \end{equation*}
    \begin{equation*}
        \text{csi} = \frac{TP}{TP + FN + FP}
    \end{equation*}
    \begin{equation*}
        \text{$\kappa$} = \frac{2 \times (TP \times TN - FP \times FN)}{(TP + FP) (FP + TN) + (TP + FN) (FN + TN)}
    \end{equation*}
    \item \textbf{AUROC/AUC}: the AUROC or AUC (Area Under the Receiver Operating Characteristic curve) represents the probability that the model correctly ranks a randomly chosen positive instance higher than a randomly chosen negative instance. The ROC curve is created by plotting the TP rate against the FP rate.
\end{itemize}

\begin{table*}[t]
\centering
\caption{Summary table containing the Label Propagation parameters}
\label{tab:LPparam}
\begin{tabular}{|l|l|l|l|l|}
\hline
Model & Kernel   & Number of neighbors & Number of iterations & Convergence threshold  \\ \hline
Label Propagation & KNN & 7 & 1000 & $10^{-3}$ \\ \hline
\end{tabular}
\end{table*}

Using these metrics, the hyperparameters of our model for the LP algorithm need to be defined. The first parameters defined are the parameters of the function used. The choice of the kernel is between KNN or RBF (radial basis function); depending on this choice, two other associated parameters could be modified. The maximum number of iterations and the algorithm's convergence tolerance remained the default values: 1000 iterations and $10^{-3}$ for the convergence threshold. The first parameter to be defined was the choice of kernel. For this, a cross-validation was carried out on the data. This involves dividing the data into several parts and then running the two algorithms using different values for the parameters on each part, keeping one part aside for performance testing. Then a calculation of the average performance can be done over all the test parts for each value and choose the one that gives the best performance. A KNN kernel with a number of neighbors of 7 has been chosen. The table \ref{tab:LPparam} summarizes the parameters chosen for the algorithm. The data are separated as follows: 70\% training and 30\% testing. The data from the training part are redivided evenly to obtain 50\% of unlabeled data and 50 \% of labeled data.

\section{Results and Discussion}
\label{sec:results}

Different proportions of the dataset were tried for annotation to optimize the Lable Propagation algorithm and achieve the best performance on automatic labeling. The aim is to annotate as few pulses as possible. 2.5\%, 5\%, 7.5\%, and 10\% of the dataset were annotated, given that the entire database contains 1571 signals, and the proportion that gave the best results was evaluated. For each proportion of the dataset, the precision, recall, and F1 values were analyzed to decide. The results for class 1 (class ''pulse with artifacts'') of these metrics are presented in the table \ref{tab:proportion}. Because the data are imbalanced, the results for class 0 (class ''pulse without artifacts'') remain consistently good and don't change much with different parameters. The best results are obtained for a proportion of 5\% of the dataset. Indeed, as the proportion of annotated data increases, the distribution of classes becomes even more disparate. For 2.5\% there are 17.3\% of pulses with artifacts, and for 5\%, the proportion of pulses with artifacts is 18.1\%. As the size of the annotated dataset increases, for 7.5\% there are 16.4\% of pulses with artifacts. For 10\% of the dataset, 17.7\% of pulses contain artifacts. All these values are summarized in the table \ref{tab:percentage_artifacts}. If the classes are more unbalanced, this may influence the algorithm, which will have greater difficulty in finding a constant pattern for propagating the labels. In the case of 10\%, the proportion of pulses with artifacts is high, but the number of annotated pulses increases, and this may induce new data that is less representative of the overall data distribution, leading to poor generalization on unseen data.  
\begin{table*}[t]
\centering
\caption{Summary table containing the labeling portion and the imbalance rate}
\label{tab:percentage_artifacts}
\begin{tabular}{|l|l|l|}
\hline
Dataset proportion    & Artifacts (\%)      & Non-artifacts (\%)  \\ \hline
2.5\% & 17.3 & 82.7 \\ \hline
5\% & 18.1 & 81.9 \\ \hline
7.5\% & 16.4 & 83.6 \\ \hline
10\% & 17.7 & 82.3\\ \hline
\end{tabular}
\end{table*}

\begin{table*}[t]
\centering
\caption{Results for the class "with artifacts" for different proportions of the dataset}
\label{tab:proportion}
\begin{tabular}{|l|l|l|l|}
\hline
Dataset proportion    & Precision      & Recall         & F1 \\ \hline
2.5\% & 0.89 & 0.88 & 0.89 \\ \hline
5\% & 0.91 & 0.90 & 0.90 \\ \hline
7.5\% & 0.84 & 0.88 & 0.86 \\ \hline
10\% & 0.83 & 0.90 & 0.86 \\ \hline
\end{tabular}
\end{table*}

\begin{table*}[t]
\centering
\caption{Results for the class "with artifacts" for different sampling methods}
\label{tab:sampling}
\begin{tabular}{|l|l|l|l|}
\hline
Sampling method    & Precision      & Recall         & F1 \\ \hline
None & 0.96 & 0.82 & 0.89 \\ \hline
RUS & 0.87 & 0.90 & 0.88 \\ \hline
ROS & 0.91 & 0.87 & 0.89 \\ \hline
SMOTE & 0.91 & 0.90 & 0.90 \\ \hline
ADASYN & 0.88 & 0.91 & 0.90 \\ \hline
ROS+RUS & 0.89 & 0.91 & 0.90 \\ \hline
\end{tabular}
\end{table*}

\begin{figure*}[t]
	\centering
	\vspace{2pt}
	\includegraphics[scale=0.9]{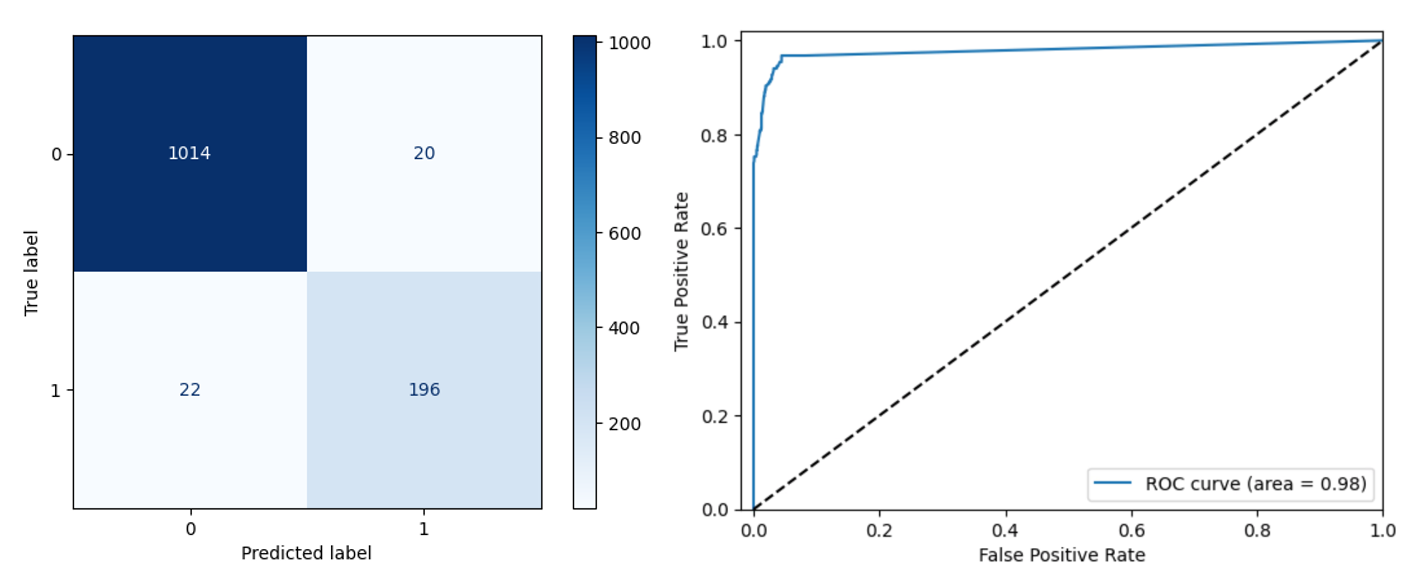}
	\vspace{-2pt}
	\caption{Confusion matrix and ROC curve for the LP algorithm with a KNN kernel with 7 neighbors, an oversampling method SMOTE, and 5\% of the dataset already labeled.}
	\vspace{-8pt}
	\label{fig:cm_roc}
\end{figure*}
The various resampling methods presented in section \ref{subsec:imbalanced} were applied. The results are shown in table \ref{tab:sampling}. First, we can see that despite the class imbalance, the algorithm manages to detect the artifacts for the training part and that the scores are correct (96\% precision, 82\% recall, and 89\% F1). However, when we apply a resampling method, the results between the scores are more balanced. This results in a more robust algorithm. The difference in results between undersampling and oversampling can be explained by the fact that undersampling will reduce the number of majority, which leads to loss of data and loss of information from this data. On the contrary, oversampling increases the number of values in the minority class, providing more data. In our case, SMOTE is the best oversampling method. SMOTE selects a minority class instance and identifies its k-nearest neighbors in the feature space. It then creates new synthetic examples along the line segments connecting the selected instance and its neighbors. By introducing these synthetic examples, SMOTE effectively increases the size of the minority class, making it comparable to the majority class and improving the performance of classifiers in handling imbalanced datasets. However, it is important to consider that resampling can potentially cause issues such as overfitting. It is important to monitor the model's performance after oversampling to detect any signs of overfitting or other potential issues. Our model showed no signs of overfitting, and resampling was very important for training a well-balanced classifier in the case of the imbalanced dataset.

\begin{table*}[t]
\centering
\caption{A comparison of the performance of different classifiers for the "with artifact" class. The oversampling method chosen is SMOTE for the LP model and ADASYN for the other models. In addition, 5\% of the dataset is already annotated.}
\label{tab:comp_class_report}
\begin{tabular}{|l|l|l|l|l|l|l|l|}
\hline
Model   & Precision      & Recall         & F1 & MCC & Kappa & CSI & AUROC \\ \hline
LP & 0.91 & 0.90 & 0.90 & 0.89 & 0.89 & 0.82 & 0.98\\ \hline
KNN & 0.89 & 0.95 & 0.92 & 0.90 & 0.90 & 0.85 & 0.98\\ \hline
MLP & 0.76 & 0.97 & 0.85 & 0.83 & 0.82 & 0.74 & 0.99\\ \hline
Transformer & 0.85 & 0.86 & 0.85 & 0.82 & 0.82 & 0.74 & 0.97\\ \hline
FCN & 0.86 & 0.83 & 0.84 & 0.82 & 0.82 & 0.74 & 0.95\\ \hline
\end{tabular}
\end{table*}

\begin{figure*}[t]
	\centering
	\vspace{2pt}
	\includegraphics[scale=0.5]{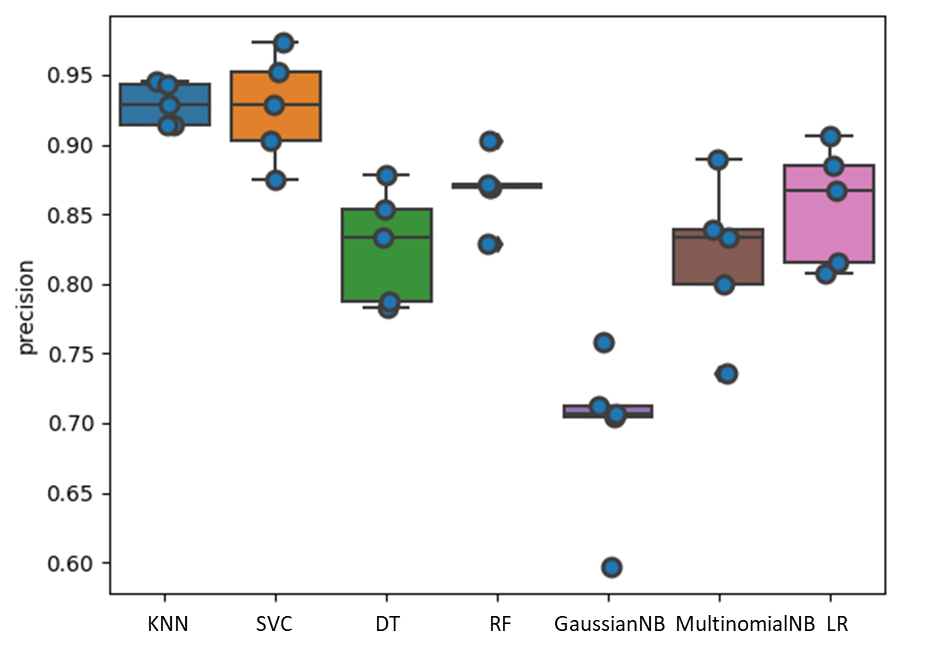}
	\vspace{-2pt}
	\caption{Precision evaluation for different traditional Machine Learning classifiers (in axis order: KNN, SVC, Decision Tree, Random Forest, Gaussian NB, Multinomial NB, Logistic Regression).}
	\vspace{-8pt}
	\label{fig:CML_compa}
\end{figure*}

\begin{figure*}[t]
	\centering
	\vspace{2pt}
	\includegraphics[scale=0.7]{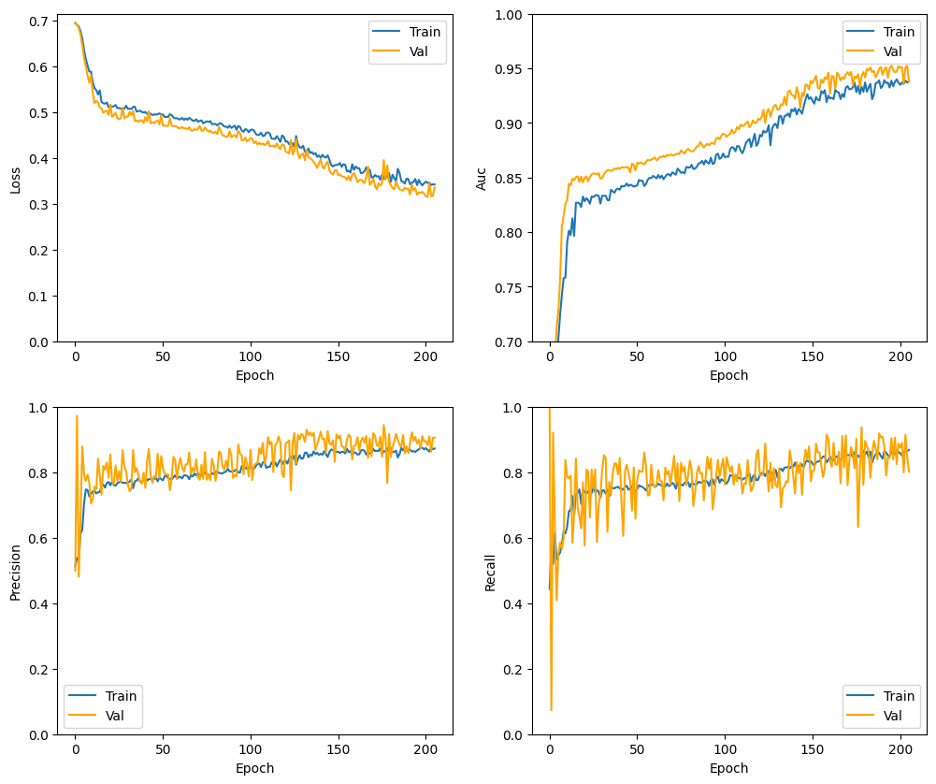}
	\vspace{-2pt}
	\caption{The comparison of CNN performance during training (Train) and validation (Val) for loss, AUROC, precision, and recall.}
	\vspace{-8pt}
	\label{fig:lossfunction}
\end{figure*}

Given the correct sampling method and the appropriate proportion of the dataset to be selected, the results of the LP algorithm were evaluated. The confusion matrix is shown on the left and the ROC curve on the right, on the Fig. \ref{fig:cm_roc}. The ROC curve is plotted for each decision threshold. In the case of the LP algorithm, this represents the probability assigned to each instance for each class. For a 5\% dataset, the number of pulses for the validation part is 1252. 1034 belong to the "without artifacts" category, and 218 belong to the "with artifacts" category. The number of true positives and true negatives is higher than the number of false positives or false negatives. This indicates that the algorithm can understand and apply the model to unlabeled signals. However, the number of pulses detected as clean but containing artifacts (false negatives) is higher than the reverse (false positives). The LP algorithm uses neighborhood information to propagate labels through the data network. This means that labels for samples close in feature space tend to be similar. However, in the case of pulses containing artifacts, these artifacts may be similar to certain features of other clean pulses, leading to incorrect label propagation. As a result, pulses containing artifacts may be incorrectly labeled as clean by the LP algorithm, leading to a higher number of falsely classified pulses. Misclassified pulses are always an important problem in the medical field. This can lead to false alarms if the pulse is not a clean pulse or to misdetections. False alarms force hospital staff to make emergency visits due to outliers. These situations are exhausting and not necessary as an additional burden on caregivers. 
Presented in Table {\ref{tab:comp_class_report}}, the three additional scores, MCC, Kappa, and CSI, support this explanation. Indeed, these three values must be close to 1 to indicate a good algorithm prediction. The closer the score is to 0, the more random the algorithm's prediction. We note that the MCC, CSI, and Kappa values demonstrate a strong prediction of the LP algorithm. These scores allow us to validate the algorithm's correct performance. 
The AUROC is 0.98. The closer the AUROC is to 1, the better the model's performance. A high AUROC indicates that our model can distinguish between positive and negative classes.

After evaluating the LP algorithm, the performance of the different types of classifiers, presented in section \ref{subsec:classification}, were assessed. First, dealing with the imbalanced classes by oversampling using the ADASYN algorithm. Cross-validation was employed to ensure accurate model prediction and assess the reliability of the machine-learning algorithms. The results of the 5-fold cross-validation on different classifiers are presented in Fig. \ref{fig:CML_compa} shows a precision comparison using a box plot. Each blue dot represents the performance of an individual fold in the cross-validation. The figure indicates that KNN and SVC (with a kernel of 'rbf') are the top-performing classifiers, with median precision rates above 90\%. However, KNN shows a slightly better and more consistent performance than SVC. This observation is highlighted by the broader range of variability for SVC, as indicated by the whiskers on the box plot, compared to KNN. To sum up, KNN and SVC are the top classifiers, but KNN is the more reliable and stable solution. KNN is also the best classifier compared to classifiers that use neural networks such as MLP and Transformers. Table \ref{tab:comp_class_report} shows the different performances of the neural networks classifiers: MLP classifier, Transformer, Fully Convolutional Network (FCN) VS KNN classifier. MLP consists of 3 hidden layers with 500 neurons for each hidden layer. Its macro average accuracy (calculates the accuracy for each class individually and then computes the average accuracy across all classes) is 0.88, compared with 0.94 for KNN. In our case, using a complex model like MLP could lead to overfitting, as the model may have a high capacity relative to the amount of data available. In addition, training an MLP can be computationally expensive, especially with larger architectures and limited computational resources. Transformers, especially large ones like BERT (Bidirectional Encoder Representations from Transformers), have a high computational complexity and require significant computational resources for training and inference. Like the MLP classifier, Transformers works best on larger datasets because it needs a lot of data for the training part. Otherwise, the model has a greater capacity than the limited data, and the risk is overfitting. Generally speaking, in the medical field, Transformers excel in natural language processing tasks \cite{yogarajan2021}. They can learn complex relationships and patterns within the text, making them suitable for medical text classification and understanding tasks.

Additionally, The LP algorithm demonstrates consistent and balanced performance in precision and recall, as evidenced by the experiment results in Fig. \ref{fig:cm_roc} and Table \ref{tab:comp_class_report}. The confusion matrix in Figure 5 shows that the LP algorithm with a KNN kernel (7 neighbors) achieves high accuracy, correctly classifying the majority of positive and negative cases, leading to a precision of 0.91 and a recall of 0.90. The ROC curve with an area of 0.98 further indicates the model's ability to distinguish between classes. Table 6 reinforces these findings by comparing the LP algorithm to other classifiers, where LP maintains competitive precision and recall values while achieving high F1 (0.90) and MCC (0.89) scores. These results highlight that the LP algorithm, combined with SMOTE for oversampling and leveraging 5\% of already labeled data, effectively balances precision and recall, ensuring robust performance in classifying the "with artifact" class.

For the last classifier, experiments were conducted with an FCN model. FCN is a neural network architecture for semantic segmentation, producing dense pixel-wise predictions. It consists of convolutional layers without fully connected layers, enabling it to handle images of any size and preserve spatial information. Using FCN for time series classification involves adapting the fully convolutional architecture to process one-dimensional time series data. Instead of working with two-dimensional images, the FCN is applied to sequences of data points. The temporal convolutional layers capture temporal patterns and dependencies in the time series, and the decoding path with transposed convolutional layers helps to produce dense predictions for each data point in the sequence, enabling accurate time series classification \cite{wang2017}. One key benefit is that FCN eliminates manual feature engineering, as they can directly learn relevant features from raw time series data. This streamlines the classification process and saves time and effort in designing handcrafted features. Additionally, FCN enables end-to-end learning, optimizing feature representations and classification jointly, which can lead to improved performance. The flexibility of FCN with input size allows them to handle time series data of varying lengths without requiring resizing or padding, making them suitable for irregularly sized data. Moreover, FCN produces dense predictions for each time step, capturing fine-grained temporal patterns and enhancing the informativeness of classification results. Experimentally, during FCN training, it is evident that the process takes longer than other approaches. However, its performance is not comparable to those methods, mainly due to its lower accuracy.

For training MLP, FCN, and Transformer, we use the binary cross-entropy loss as follows: 
\begin{equation}
    L_{BCE}=-\frac{1}{n}\sum_{i=1}^{n}(Y_i \times log(\hat{Y}_i)+(1-Y_i) \times log(1-\hat{Y}_i))
\end{equation}

We use the Adam optimizer and early stopping to deal with the overfitting. We use GridSearchCV to fine-tune the hyper-parameters, balancing the best combination and computation time. Only certain hyper-parameters typically affect a neural network's accuracy, specifically the number of hidden layers, nodes in each hidden layer, and the learning rate \cite{luo2016review}. By focusing on these, grid search effectively optimizes all parameters simultaneously, allowing for quick model training. Grid search also offers straightforward parallelization and flexible resource allocation, which other approaches lack \cite{yu2020hyper}.

Fig. \ref{fig:lossfunction} provides a comprehensive view of the model's performance over time. The improvements in metrics like loss, AUC, precision, and recall suggest that the model is learning and improving its performance with each epoch. The consistent trends between training and validation data indicate that the model is generalizing well and not overfitting significantly. However, we can see the fluctuation between precision and recall; it can be confirmed that FCN can not deal with the imbalanced classes from the nature of the data.

So, compared with previous studies, the artifact classification algorithm we have implemented has the advantage of having a faster execution time on large volumes of data compared to EMD or wavelet denoising, for example. It exploits the intrinsic relationships between the data rather than decomposing each signal individually. It also has the advantage of being easily generalizable to other signals since no additional parameters are required. 

\section{Limitations and future work}
This study delved into utilizing semi-supervised LP methods for artifact classification within PPG signals, especially in scenarios characterized by imbalanced class distributions. The study showed us that our model is sensitive to data volume, and its improvement is limited as data volume increases. One future objective is to improve our model, particularly in feature detection. To augment the capability of our model, we can add some steps in the preprocessing part. In section \ref{subsec:stat}, the segmentation problem has already been mentioned. First, adaptive filtering techniques can attenuate artifacts without affecting the signal. Signal quality can be improved through noise reduction methods, such as singular value decomposition (SVD). Alternative segmentation approaches can be employed to enhance the efficiency of the statistical analysis algorithm. Peak or minimum detection can be improved by employing derivative-based algorithms. A CNN model can also detect peaks, known for its pattern recognition capabilities \cite{kazemi2022}. Implementing alternative segmentation approaches and employing a CNN model for peak detection in PPG signals may face challenges in parameter tuning, validation, and network architecture design. However, careful optimization and validation of these approaches can improve the algorithm's accuracy and reliability.

Exploring data augmentation can also be a method to tackle the model's sensitivity to data volume. The authors in \cite{khan2023} found that using data augmentation allowed them to better handle the unbalanced class problem for binary or multiclass classification. These authors also draw a parallel with ensemble learning methods, which are new hybrid methods that are more robust against unbalanced data. In \cite{shen2022, zandbagleh2023}, the authors put into practice the use of RUSBoost for epilepsy seizure detection and schizotypy classification. Compared to classical models, the ensemble model performed very well, making it a candidate for feature classification. Investigating ensemble learning models may help to better handle class imbalances and classification tasks. Exploring self-supervised or unsupervised learning methods could be considered for future work to address the labeling challenge we encountered without relying on manual annotations.

\section{Conclusion}
\label{sec:conclusion}
This study explored applying semi-supervised  LP methods for artifact classification in PPG signals, addressing challenges posed by imbalanced class distributions. Comparative analysis with traditional supervised learning algorithms, MLP, and Transformer-based models demonstrated the superior performance of the LP classifier. 
This algorithm can enhance the overall quality of PPG signals in artifact classification by dynamically adapting to the specific characteristics of the dataset. It becomes more adaptable to variations in PPG signals caused by different types of motion artifacts. The improved balance between precision and recall indicates more robust classifier performance, which is critical for real-life medical applications.

Overall, this model holds promise for enhancing healthcare monitoring systems, with potential applications in ECG and arterial blood pressure signal analysis.

\section*{Acknowledgment}
The clinical data used in this study were generously provided by the Research Center of the CHU Sainte-Justine Hospital, University of Montreal. The authors thank Dr. Sally Al Omar, Dr. Michael Sauthier, and Edem Tiassou for their data support of this research and K\'evin Albert for annotating the data. 

\bibliographystyle{IEEEtran}
\bibliography{IEEEabrv,Bibliography}

\vspace{-12em}

\begin{IEEEbiography}[{\includegraphics[width=1in,height=1.25in,clip,keepaspectratio]{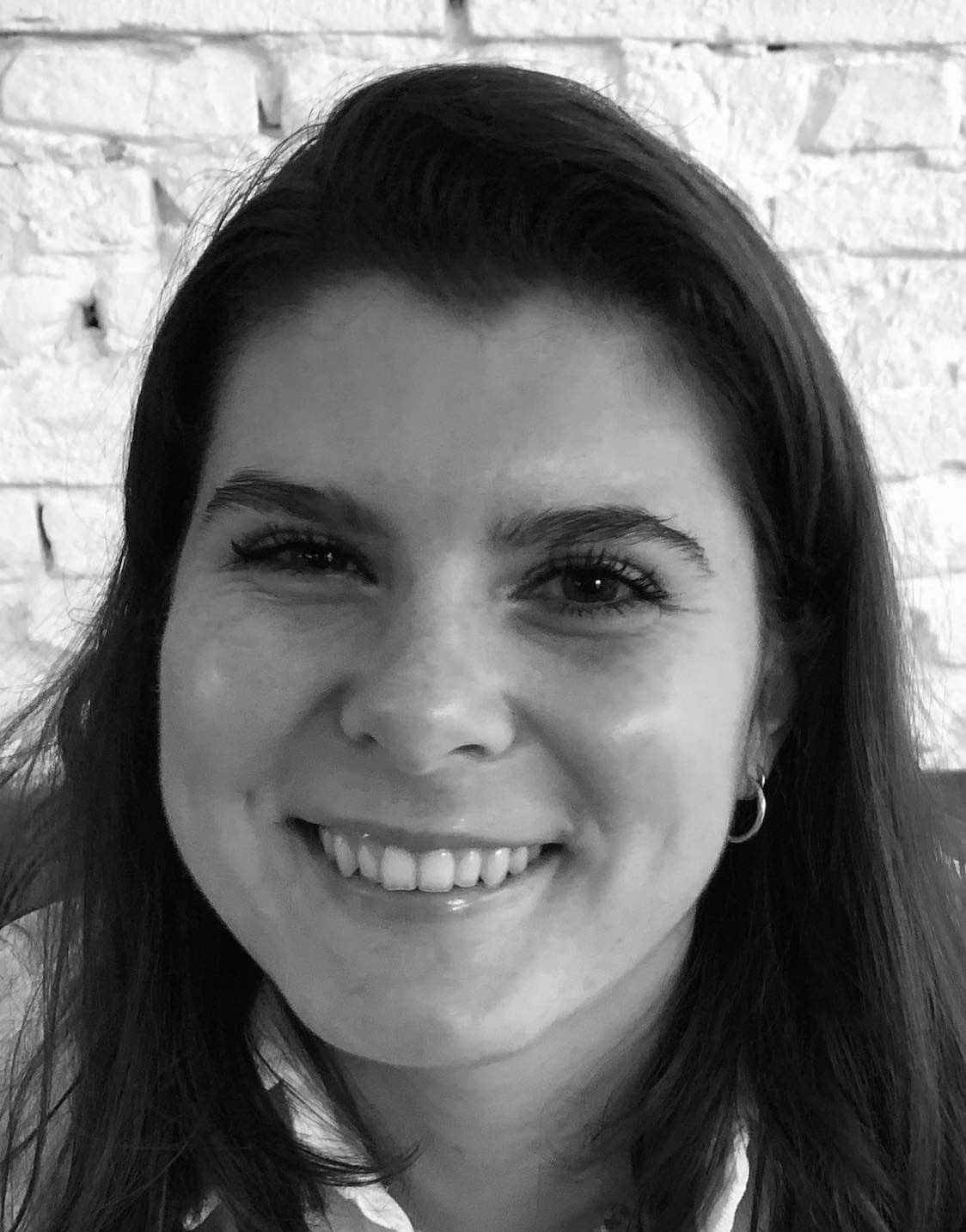}}]{Clara Macabiau} is a double degree student in Canada. After three years at the ENSEEIHT engineering school in Toulouse, specializing in EEEA (Electronics, Electrical Energy and Automation), she is completing her master's degree with a thesis in electrical engineering at the Ecole de Technologie Supérieure in Montreal. Her master's project focused on the detection of artifacts in photoplethysmography signals from children admitted to pediatric intensive care at CHU Sainte-Justine. Her fields of interest are signal processing, machine learning and electronics.
\end{IEEEbiography}

\vskip -2\baselineskip plus -1fil

\begin{IEEEbiography}[{\includegraphics[width=1in,height=1.25in,clip,keepaspectratio]{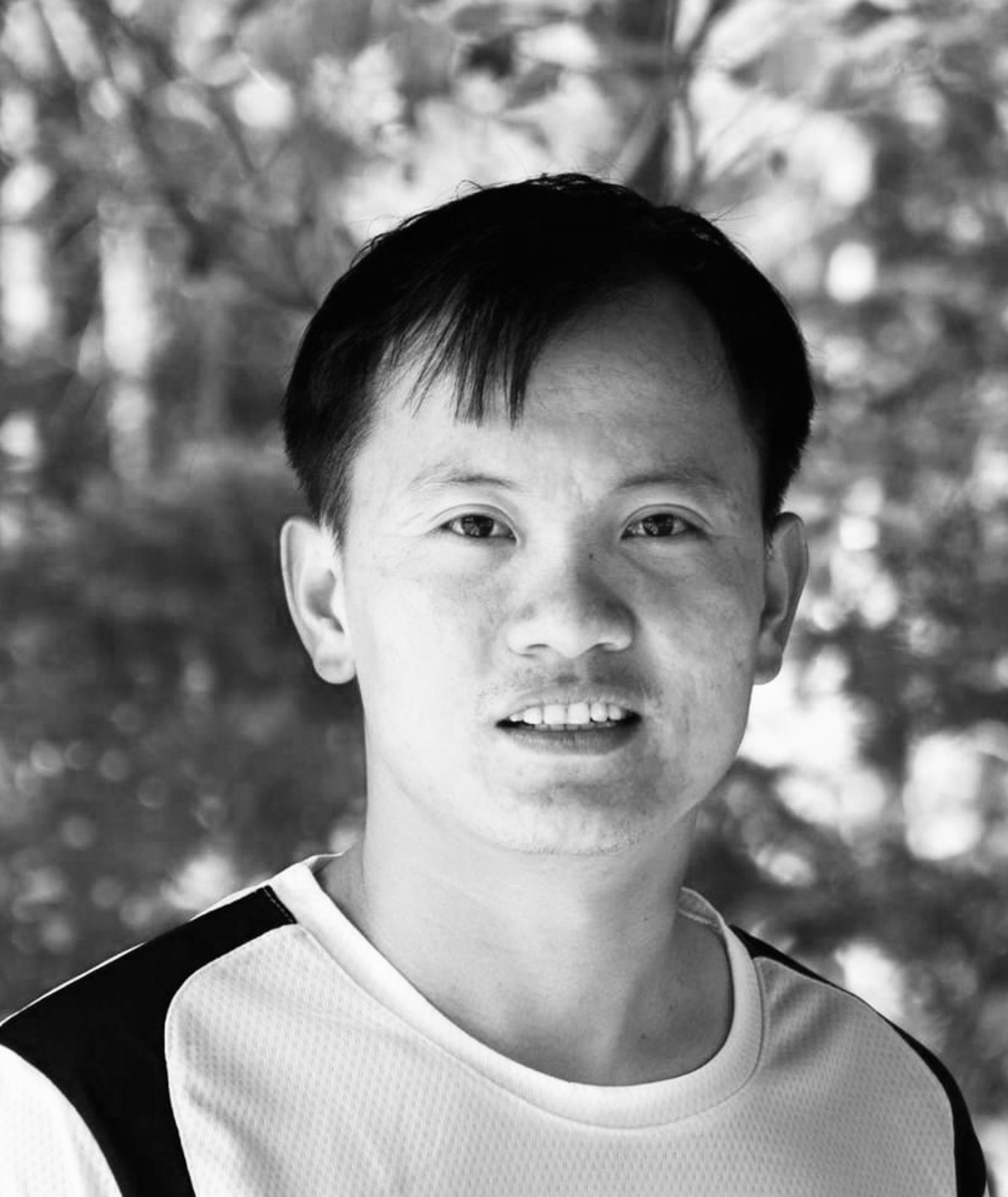}}]{Thanh-Dung Le} (Member, IEEE) received a B.Eng. degree in mechatronics engineering from Can Tho University, Vietnam, an M.Eng. degree in electrical engineering from Jeju National University, S. Korea, and a Ph.D. in biomedical engineering from Ecole de Technologie Supérieure (ETS), Canada. He is a postdoctoral fellow at the Biomedical Information Processing Laboratory, ETS. His research interests include applied machine learning approaches for biomedical informatics problems. Before that, he joined the Institut National de la Recherche Scientifique, Canada, where he researched classification theory and machine learning with healthcare applications. He received the merit doctoral scholarship from Le Fonds de Recherche du Quebec Nature et Technologies. He also received the NSERC-PERSWADE fellowship,  Canada, and a graduate scholarship from the Korean National Research Foundation, S. Korea.
\end{IEEEbiography}

\vskip -2\baselineskip plus -1fil

\begin{IEEEbiography}[{\includegraphics[width=1in,height=1.25in,clip,keepaspectratio]{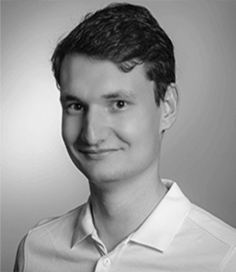}}]{Kévin Albert} is physiotherapist, graduated from EUSES School of Health and Sport (2018 - Girona, Spain). He developed clinical expertise in the field of function rehabilitation after neuro-traumatic injury (France) and in cardio-respiratory rehabilitation (Swiss). He is currently enrolled in the Master's Biomedical Engineering program at the University of Montreal and has joined the Clinical Decision Support System (CDSS) laboratory under the supervision of Prof. P. Jouvet M.D. Ph.D. in the Pediatric Intensive Care Unit at Sainte-Justine Hospital (Montréal, Canada) since May 2023. His primary research interest is application new technologies of support care system tool with artificial intelligence, especially in ventilatory support. His research program is supported by the Sainte-Justine Hospital and the Quebec Respiratory Health research Network (QRHN).
\end{IEEEbiography}

\vskip -2\baselineskip plus -1fil

\begin{IEEEbiography}[{\includegraphics[width=1in,height=1.25in,clip,keepaspectratio]{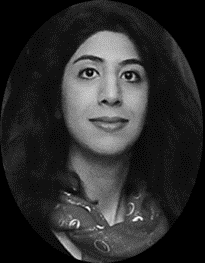}}]{Mana Shahriari} is an artificial intelligence (AI) researcher passionate about employing AI to address practical and real-world challenges. Her research interests are signal processing (including time-series analysis), image processing and computer vision, machine learning and deep learning, and statistical analysis of data. She is currently a postdoctoral researcher at CHU Sainte Justine
research centre, affiliated with University of Montreal. She holds a Ph.D. in electrical engineering, a Master's in Artificial Intelligence, and a bachelor’s in electrical engineering.
\end{IEEEbiography}

\vskip -2\baselineskip plus -1fil

\begin{IEEEbiography}[{\includegraphics[width=1in,height=1.25in,clip,keepaspectratio]{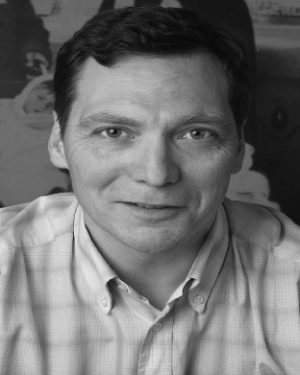}}]{Philippe Jouvet} received the M.D. degree from Paris V University, Paris, France, in 1989, the M.D. specialty in pediatrics and the M.D. subspecialty in intensive care from Paris V University, in 1989 and 1990, respectively, and the Ph.D. degree in pathophysiology of human nutrition and metabolism from Paris VII University, Paris, in 2001. He joined the Pediatric Intensive Care Unit of Sainte Justine Hospital—University of Montreal, Montreal, QC, Canada, in 2004. He is currently the Deputy Director of the Research Center and the Scientific Director of the Health Technology Assessment Unit, Sainte Justine Hospital–University of Montreal. He has a salary award for research from the Quebec Public Research Agency (FRQS). He currently conducts a research program on computerized decision support systems for health providers. His research program is supported by several grants from the Sainte-Justine Hospital, Quebec Ministry of Health, the FRQS, the Canadian Institutes of Health Research (CIHR), and the Natural Sciences and Engineering Research Council (NSERC). He has published more than 160 articles in peer-reviewed journals. Dr. Jouvet gave more than 120 lectures in national and international congresses.
\end{IEEEbiography}

\vskip -2\baselineskip plus -1fil

\begin{IEEEbiography}[{\includegraphics[width=1in,height=1.25in,clip,keepaspectratio]{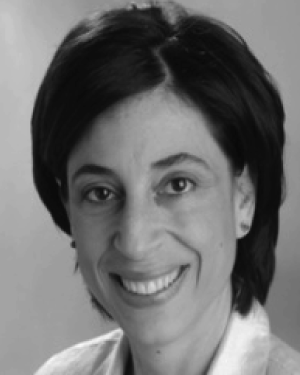}}]
{Rita Noumeir} (Member, IEEE) received master's and Ph.D. degrees in biomedical engineering from École Polytechnique of Montreal. She is currently a Full Professor with the Department of Electrical Engineering, École de Technologie Superieure (ETS), Montreal. Her main research interest is in applying artificial intelligence methods to create decision support systems. She has extensively worked in healthcare information technology and image processing. She has also provided consulting services in large-scale software architecture, healthcare interoperability, workflow analysis, and technology assessment for several international software and medical companies, including Canada Health Infoway.
\end{IEEEbiography}

\EOD

\end{document}